\documentclass{article}

\usepackage{arxiv}

\usepackage{amssymb}
\usepackage{rotating}
\usepackage{floatrow}
\usepackage{multirow}
\usepackage{graphicx}
\usepackage{soul}
\usepackage{wrapfig}
\usepackage{hyperref}

\hypersetup{
    colorlinks=true,
    linkcolor=blue,
    filecolor=blue,      
    urlcolor=blue,
    }

\usepackage{comment}

\usepackage{multicol}
\usepackage{comment}
\usepackage{booktabs, makecell, tabularx}
\usepackage{xspace}
\usepackage{amsfonts}
\usepackage[dvipsnames,table,xcdraw]{xcolor}
\definecolor{gold}{rgb}{0.85, 0.65, 0.13}
\usepackage{algorithm}
\usepackage[noend]{algpseudocode}

\algnewcommand\algorithmicforeach{\textbf{for each}}
\algdef{S}[FOR]{ForEach}[1]{\algorithmicforeach\ #1\ \algorithmicdo}
\algnewcommand{\algorithmicand}{\textbf{ and }}
\algnewcommand{\algorithmicor}{\textbf{ or }}
\algnewcommand{\OR}{\algorithmicor}
\algnewcommand{\AND}{\algorithmicand}
\algnewcommand{\LineComment}[1]{\State \(\triangleright\) {\footnotesize #1}}
\newcommand{\ea}{\textit{et. al.}\xspace}
\newcommand{\method}{CANTS-N\xspace}

\algdef{SE}[VARIABLES]{Variables}{EndVariables}
   {\algorithmicvariables}
   {\algorithmicend\ \algorithmicvariables}
\algnewcommand{\algorithmicvariables}{\textbf{global variables}}

\usepackage[caption=false]{subfig}
\usepackage{pgfplots}

\pgfplotsset{compat=1.12}

\newcommand{\todo}[1]{{\color{purple}\textbf{\ul{TODO}:} \emph{#1}}}

\title{Colony-Enhanced Recurrent Neural Architecture Search: Collaborative Ant-Based Optimization}

\author{ 
    AbdElRahman ElSaid
	\thanks{Department of Computer Science, University of North Carolina Wilmington}\\
	\texttt{elsaida@uncw.edu} \\
}

\renewcommand{\shorttitle}{CANTS-N}

\hypersetup{
pdftitle={A template for the arxiv style},
pdfsubject={q-bio.NC, q-bio.QM},
pdfauthor={David S.~Hippocampus, Elias D.~Striatum},
pdfkeywords={First keyword, Second keyword, More},
}

\begin{document}
\maketitle

\begin{abstract}
    Crafting neural network architectures manually is a formidable challenge, often leading to suboptimal and inefficient structures. The pursuit of the perfect neural configuration is a complex task, prompting the need for a metaheuristic approach such as Neural Architecture Search (NAS). Drawing inspiration from the ingenious mechanisms of nature, this paper introduces Collaborative Ant-based Neural Topology Search (CANTS-N), pushing the boundaries of NAS and Neural Evolution (NE). In this innovative approach, ant-inspired agents meticulously construct neural network structures, dynamically adapting within a dynamic environment, much like their natural counterparts. Guided by Particle Swarm Optimization (PSO), CANTS-N's colonies optimize architecture searches, achieving remarkable improvements in mean squared error (MSE) over established methods, including BP-free CANTS, BP CANTS, and ANTS. Scalable, adaptable, and forward-looking, CANTS-N has the potential to reshape the landscape of NAS and NE. This paper provides detailed insights into its methodology, results, and far-reaching implications.
\end{abstract}




\section{Introduction}
\label{sec:intro}

In the realm of nature, ant colonies frequently showcase specialized roles within their communities. For instance, some colonies may dedicate most of their effort to foraging for food, while others prioritize the defense of their nests or the nurturing of their brood. Intriguingly, these colony-specific behaviors can undergo transformations throughout the colony's lifespan, influenced by interactions with the environment and neighboring colonies~\cite{gordon2010ant}.

Translating this concept into the domain of Neural Architecture Search (NAS) yields fascinating possibilities. Just as diverse ant colonies coexist in the environment, each with a unique focus, multi-colony NAS can manifest as a specialized community of colonies, each with a distinct role within the NAS task. Some colonies may excel at exploring a broad spectrum of network architectures, while others might demonstrate proficiency in exploiting promising designs. This diversity contributes to an efficient coverage of the search space.

In the natural world, multiple ant colonies simultaneously explore various regions of their habitat. Similarly, multi-colony NAS enables parallel exploration of the neural architecture search space. This approach fosters comprehensive exploration, enhancing the likelihood of uncovering optimal architectures. Just as nature's ant colonies compete for resources, NAS colonies compete in the quest for superior neural architectures. Yet, akin to their natural counterparts, NAS colonies also exhibit cooperative tendencies when facing common challenges. In NAS colonies, while competition exists to some degree, the colonies ultimately collaborate through information sharing to collectively optimize the NAS process. This interplay between competition and cooperation engenders a dynamic optimization process.

Moreover, ant colonies in the natural world exhibit adaptive behaviors in response to changing environmental conditions. In NAS colonies, each colony can tailor its strategies not only based on its own experiences but also by considering shared information from other colonies. This adaptability contributes to an overall enhancement in the optimization process's performance.

Ant-based Neural Topology Search (ANTS) and Continuous Ant-based Neural Topology Search (CANTS) are established methodologies in the realm of NAS and NeuroEvolution~\cite{elsaid2017optimizing,elsaid2018optimizing,elsaid2019ant,elsaid2020ant,elsaid2021continuous,elsaid2023backpropagation}. These methods have traditionally featured a singular optimization colony.

However, in this paper, we elevate the landscape by introducing the innovative concept of multiple parallel colonies. These colonies harmoniously coexist, operate, engage in healthy competition, and foster collaboration. They unite in the pursuit of uncovering the summit of neural architecture—an innovative and promising endeavor.

Our investigation delves into the remarkable potential of multiple colonies, revealing their capacity to significantly enhance the efficiency and effectiveness of the NAS process. This synergy among colonies opens new horizons in the quest for optimal neural architectures.

In this work, we introduce Collaborative Ant-based Neural Topology Search (CANTS-N), a novel approach that leverages the power of ant colonies and Particle Swarm Optimization (PSO) to dynamically evolve Recurrent Neural Network (RNN) topologies. CANTS-N's audacious approach to RNN architecture optimization leads to state-of-the-art results, outperforming established methods by an order of magnitude.

The remainder of this paper is organized as follows:
\textbf{Section~\ref{sec:litrature}: Related Work} explores the existing literature and discusses relevant research in the field.
\textbf{Section~\ref{sec:method}: Methodology} provides a brief overview of prior work and outlines the methodology employed in the development of our method, which we will subsequently name.
\textbf{Section~\ref{sec:result}: Experimental Results} presents the experiments conducted with our method, showcasing the outcomes and performance metrics.
\textbf{Section~\ref{sec:summary}: Summary and Future Directions} offers a summary of the obtained results and outlines potential avenues for future research stemming from this project.

We will now delve into each of these sections to provide a comprehensive understanding of our work.

\section{Related Work}
\label{sec:litrature}

Lankford~\ea~\cite{lankford2020neural} explored the application of Ant Colony Optimization (ACO)~\cite{socha2008ant, manfrin2006parallel, bianchi2002ant, dorigo1997ant, dorigo1997ant_} and Particle Swarm Optimization (PSO) in OpenNAS, focusing on optimizing Convolutional Neural Network (CNN) hyperparameters and layer types. Their work laid the foundation for harnessing nature-inspired algorithms in the domain of neural architecture search.

Deep Swarm, proposed by Byla~\ea~\cite{byla2020deepswarm}, introduced a Colony System that defined nodes at various neural architecture layers, allowing agents to select layer types during the evolution process. By leveraging ACO, Deep Swarm provided a novel approach to evolve CNN structures.

Another avenue of research delved into optimizing deep neural networks using PSO. Linchao~\ea~employed PSO with a multi-objective function, incorporating mean absolute error, mean absolute percentage error, and root mean squared error. Their method, Multi-Objective Particle Swarm Optimization (MOPSO)~\cite{coello2004mopso}, aimed to enhance the stability of deep belief networks (DBNs) for traffic delay forecasting. Results showed that MOPSO outperformed single-objective PSO-optimized DBNs and non-optimized structures, highlighting the effectiveness of multi-objective PSO optimization.

Wu~\ea~\cite{wu2019multi} also utilized MOPSO but with a focus on pruning CNN structures. Unlike traditional NAS methods, this approach fixed the search space to a specific initial structure, resulting in sparser yet more efficient architectures.

In the context of CNN architecture, Wang~\ea~\cite{wang2019evolving} introduced a method that sought to strike a balance between classification accuracy and inference latency. Their approach, Multi-Objective CNN (MOCNN), leveraged OMOPSO~\cite{sierra2005improving} to optimize CNN architectures, showcasing the versatility of PSO-based techniques.

Jiang~\ea~\cite{jiang2020efficient} proposed a Multi-Objective Particle Swarm Optimization with Dominance (MOPSO/D) algorithm that featured an adaptive penalty-based boundary intersection. This adaptive algorithm was applied to optimize CNN structures, highlighting the applicability of PSO in neural architecture search.

While the literature abounds with research on NAS and NE applied to CNN architecture, a significant gap remains in the domain of RNN topology optimization. As revealed by Liu~\ea~\cite{liu2021survey} in their comprehensive survey, investigations into RNN topology optimizations are notably scarce. Addressing this research gap underscores the significance of our work, which contributes to bridging this void.

Moreover, the majority of existing NAS and NE methods primarily rely on genetic algorithms (GA)~\cite{liu2021survey}. While GAs have demonstrated success in optimizing neural architectures, they are susceptible to local minima, as their search often commences from minimal structural elements, progressively expanding the architecture through evolutionary processes\cite{elsaid2020ant}.

Unlike many NAS and NE techniques, the Continuous Ant-based Neural Topology Search (CANTS) method, as introduced 
the literature~\cite{elsaid2021continuous,elsaid2023backpropagation,elsaid2024softwareimpacts}, transcends the limitations of fixed architectures. CANTS innovatively explores an unbounded search space, enabling the generation of architectures through an evolutionary process.

\section{Methodology}
\label{sec:method}
Our methodology combines Continuous Ant-based Neural Topology Search (CANTS) with Collaborative Ant-based Neural Topology Search (CANTS-N). These techniques utilize colonies of agents guided by Particle Swarm Optimization (PSO) to explore the neural architecture search space. The following describes our approach, highlighting key components and their roles in efficient NAS.

\subsection{Continuous Ant-based Neural Topology Search (CANTS)}
\begin{figure*}
    \centering
    \includegraphics[width=.7\textwidth]{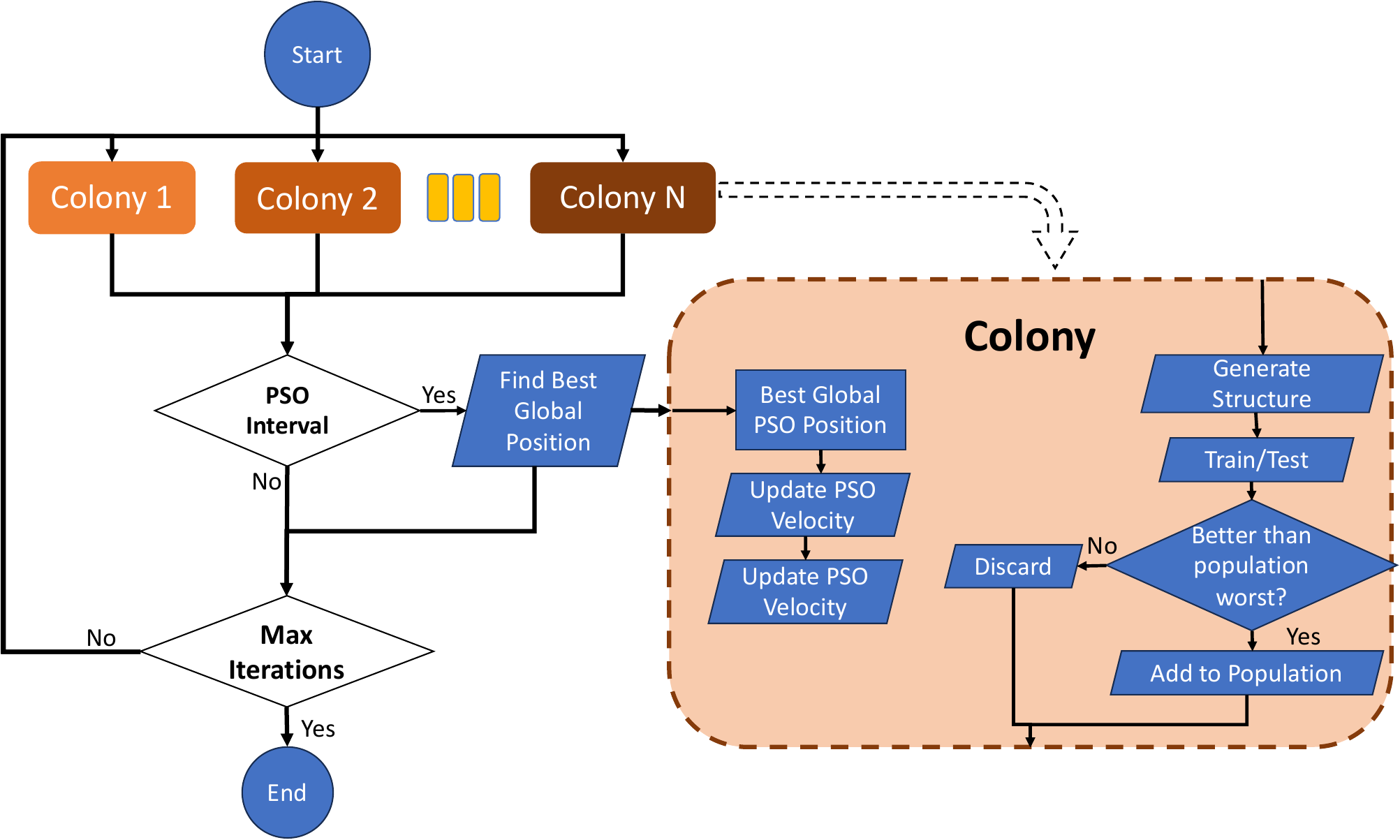}
    \caption{\textit{\textbf{Multi-Colony Flowchart}}}
    \label{fig:multi_col_flowchart}
\end{figure*}
CANTS can be likened to an ant colony, where its agents play the role of diligent foraging ants. Much like how ants gather food to sustain their colony, these agents contribute by generating neural network structures. The 'food' in this analogy represents the fitness of the structures they construct as they traverse from input nodes to output nodes. These agents engage in interactions, not only amongst themselves but also with the dynamic environment – the search space. This dynamic interaction and adaptation are akin to the thriving spirit of a vibrant colony.

\begin{figure}
    \centering
    \includegraphics[width=.48\textwidth]{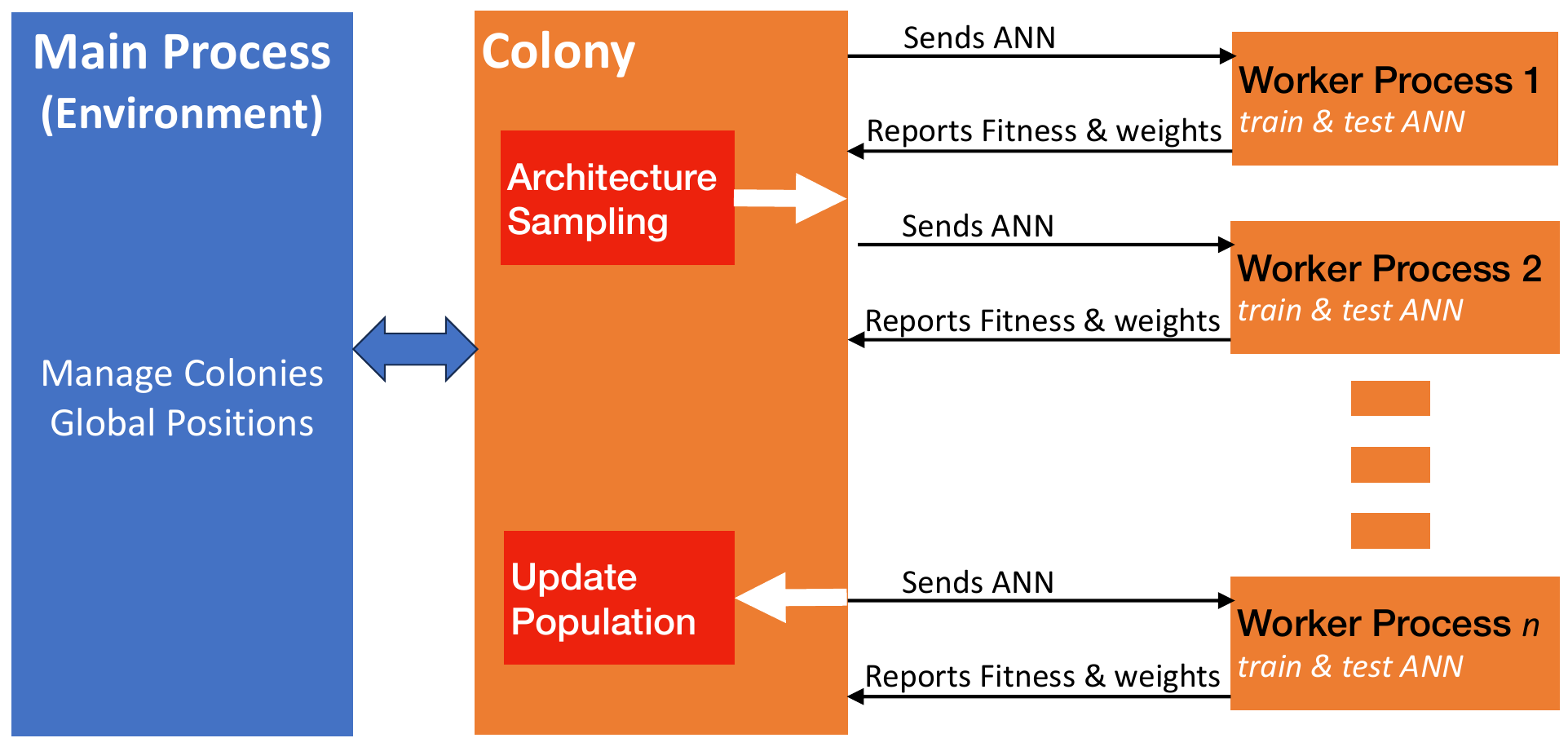}
    \caption{\textit{\textbf{Asynchronous Colonies}}}
    \label{fig:async_colonies}
\end{figure}

The CANTS procedure, as outlined in Figure~\ref{fig:async_colonies}, employs an advanced asynchronous and distributed 'work-stealing' strategy tailored for efficient execution on High-Performance Computing (HPC) systems. This dynamic approach consists of a primary process responsible for generating neural network structures sampled from the 3D continuous search space. Additionally, there are worker processes dedicated to training and validating these neural structures using training and validation data. Subsequently, these worker processes report the fitness of the structures back to the main process. This distributed work distribution significantly accelerates the search process and enhances scalability.

This design allows workers to train the generated RNNs at their own pace, resulting in a naturally balanced workload distribution. In contrast to traditional synchronous parallel evolutionary strategies, CANTS seamlessly scales to accommodate any number of available processors, making it adaptable to diverse computational environments. Furthermore, it supports population sizes that are independent of processor availability.

When a candidate RNN reports its fitness, measured as mean squared error over validation data, to the work generator process, it undergoes a rigorous evaluation. If the candidate RNN outperforms the weakest RNN in the population, the weakest RNN is replaced with the candidate. Notably, this replacement process also involves an increment in the saved pheromone placement points within the continuous search space.

CANTS successfully indirectly maps the discrete neural graph search space to a unbounded continuous 3D search space. In addition, it's noteworthy that the individual ants (agents) within the system exhibit self-awareness and evolve throughout the process, gradually becoming more adept at navigating the search space and improving their fitness for the task.

The authors refers the reader to these literature publications~\cite{elsaid2021continuous,elsaid2023backpropagation} for further details and illustrations on the CANTS procedure and its associated methodologies.

\subsection{Collaborative Ant-based Neural Topology Search (CANTS-N)}

In our method, we harness the power of CANTS (Continuous Ant-based Neural Topology Search) to establish and manage individual colonies, where each colony operates as a CANTS-N colony. These colonies work synergistically within a framework guided by PSO. The initialization of each colony is a stochastic process, resulting in unique PSO positions based on factors such as the number of ants, pheromone evaporation rate, and ants mortality rate. Additionally, each colony begins with a specific number of foraging ants.

\subsubsection{Inter-colonial communication}

\begin{algorithm}
\caption{Abstract-Level Pseudo-code: Asynchronous Multi-Ant Colony Algorithm with PSO for ANN Optimization\label{alg:multi_col}}
\begin{algorithmic}[1]
\State Initialize parameters:
\State $num\_colonies = N$  \Comment{Number of ant colonies}
\State $num\_iterations = I$  \Comment{Number of optimization iterations}
\State $pso\_interval = 20$  \Comment{Interval for PSO optimization}

\State Initialize ANN parameters:

\State Initialize colonies and their parameters:
\For{$colony\_id$ \textbf{in} $range(num\_colonies)$}
    \State Initialize colony-specific parameters:
    \State Initialize $n: colony.num\_ants$
    \State Initialize $\rho: colony.pheromone\_evaporation\_rate$
    \State Initialize $\mu: colony.ant\_mortality\_rate$
    \State $particle\_pos, particle\_vel \gets InitializePSO(n, \rho, \mu)$
\EndFor

\State Assign environment role to a process
\State Assign each colony to a process
\State Assign each colony-worker to a process

\State Main optimization loop (environment):
\For{$iteration$ \textbf{in} $num\_iterations$}
    \ForEach {$colony$}
        \State Generate a complete ANN architecture using ant
        \State Evaluate architecture fitness asynchronously
        \State Update pheromone, evaporation, and mortality
    \EndFor
    \If{$iteration$ \% $pso\_interval \gets 0$}
        \State Apply PSO optimization for colony
        \State Broadcast global best values of num\_ants, evap\_rate, and mortality\_rate
    \EndIf
\EndFor
\end{algorithmic}
\end{algorithm}

Throughout the colony's lifespan with CANTS-N, the number of foraging ants adeptly changes based on their own experiences and shared knowledge from coexisting colonies within CANTS-N. The evaporation rate determines how quickly the pheromone deposited by ants diminishes over time, making less attractive paths less appealing for new foraging ants when generating new RNNs. The mortality rate governs the rate at which ants perish, allowing new foraging ants to be born with different foraging behaviors, thereby increasing exploration opportunities for discovering better RNN structures. This dynamic colony behavior closely mirrors the adaptability seen in natural ant colonies, where foraging and scouting activities change over time in response to colony experiences and ecosystem dynamics \cite{gordon2010ant}.

These colonies continue to evolve throughout the Neural Architecture Search (NAS) optimization process, marked by the sharing of their best individual positions and learning from the best global positions achieved by any collaborating colony. In this dynamic process, each colony adjusts its PSO velocity and, consequently, its position, contributing to the ongoing optimization of the NAS task.

For a visual representation of this intricate process, please consult Figure~1, which illustrates the step-by-step evolution of neural structures as colonies generate new topologies using the ant-based evolutionary strategy, CANTS \cite{elsaid2021continuous}, over a predetermined number of optimization iterations. Moreover, at specific intervals, each colony communicates its local best position to other colonies, and the best global position among them is shared, leading to collective position adjustments across all colonies.

Algorithm~\ref{alg:multi_col} provides pseudo code illustrating how the colonies are initialized and how they communicate via PSO to evolve through optimization iterations in the NAS process at specific intervals.

\subsubsection{Asynchronous Collaborative Optimization}

As part of our approach, we leverage CANTS' inherent workload distribution between main and worker processes, introducing an additional process that serves as a shared environment for these coexisting ant colonies. Each colony maintains its dedicated main process and worker processes, facilitating asynchronous collaboration.

This collaborative paradigm functions akin to a synchronized ensemble, where each colony assumes a distinct role in harmony. The colony's main process seamlessly communicates with our method's primary process, ensuring efficient information exchange and coordination. A visual representation of this process is provided in Figure 2. Inter-process communication is facilitated through the utilization of the Message Passing Interface (MPI)~\cite{mpi_1, mpi2,mpi3, mpi4}. 

The code for the CANTS-N is implemented in Python, employing MPI, PyTorch (only for backpropagation graph creation), and NumPy, is available online. For access to the code repository, please refer to the footnote~\footnote{The URL of the GitHub repository for the CANTS-N implementation will be provided upon acceptance of the paper, in accordance with double-blind review requirements.} below.

\begin{figure*}[h]
    \centering
    \includegraphics[width=.95\textwidth, height=.4\textwidth]{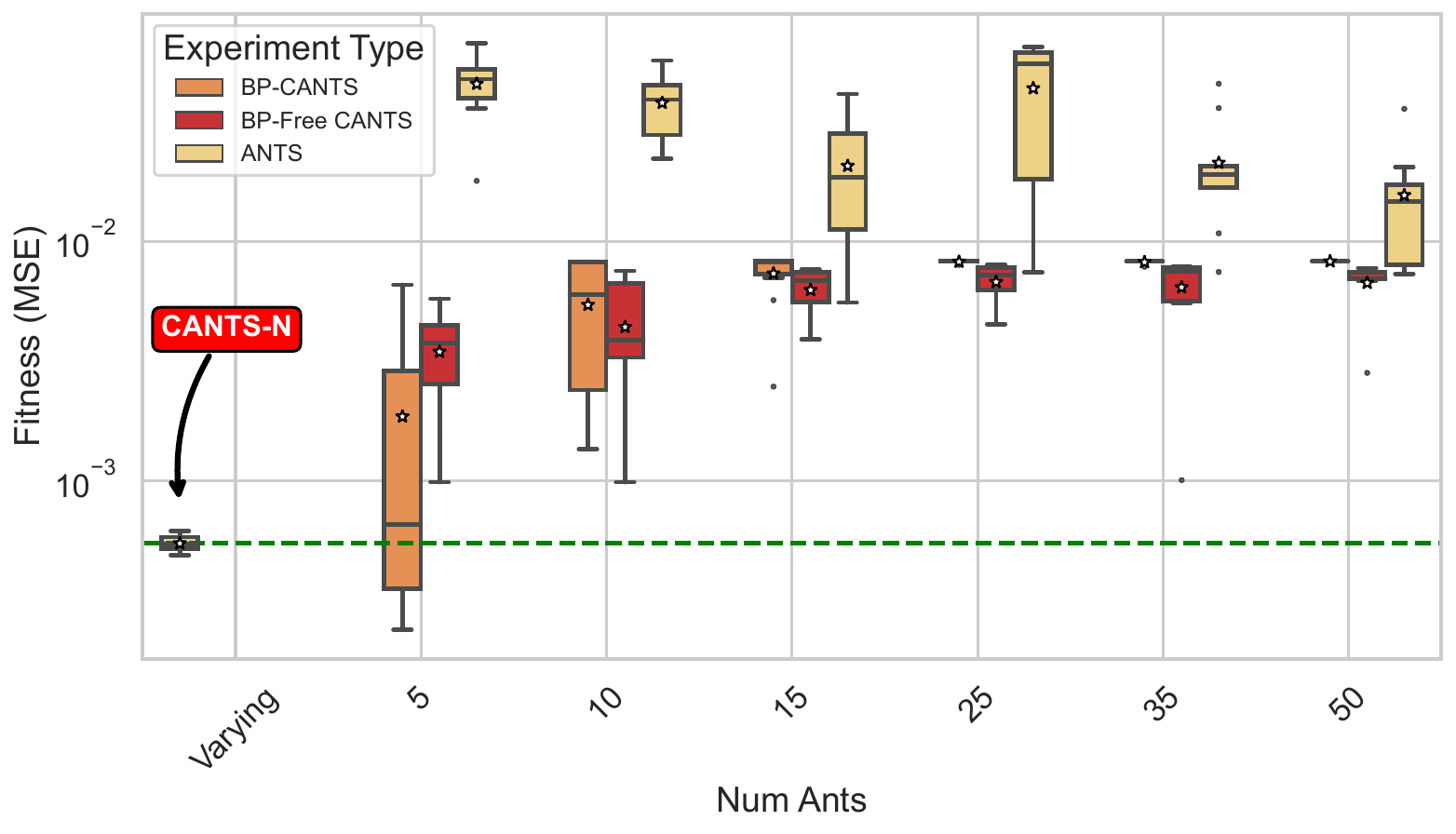}
    \caption{\centering\textit{\textbf{CANTS-N Fitness:} comparing CANTS-N (multi-colony) performance using the MSE of its colonies' best performing neural structures, to the best performing neural networks obtained from BP-CANTS, BP-free CANTS, and ANTS}}
    \label{fig:fit_box}
\end{figure*}

\section{Experiments \& Results}
\label{sec:result}

The experiment utilized 201 CPU cores on a high-performance computing (HPC) cluster, consuming approximately 80GB of RAM memory. Each core ran a separate MPI process, and the experiment was extended for a duration of 4 days. 

Twenty colonies were created, each generating 1000 RNN structures. These colonies exchanged their parameters (number of ants, evaporation rate, and ant mortality rate) every 9 RNN generations. 
One process acted as the environment, collecting the PSO best colonies' local positions and broadcasting the best global PSO position to all the colonies. Each colony ran as a separate process, managing worker processes. Nine worker processes were allocated for each colony, responsible for training and validating the RNN structures generated by their respective colonies, thus accelerating the process.
The initialization of colonies involved assigning a number of ants to each colony, ranging from 5 for the first colony to 100 for the twentieth colony. Additionally, the pheromone evaporation and mortality rates for the ants in all colonies were initially set to 0.9 and 0.1, respectively. During the PSO optimization process, the colonies were allowed to have a number of ants ranging from 10 to 200, an evaporation rate between 0.15 and 0.95, and an ants mortality rate between 0.01 and 0.1. These boundaries were defined to ensure the parameters remained within realistic values for the optimization process.

Results were compared with those obtained from BP-free CANTS, BP-CANTS, and ANTS, as reported in the literature~\cite{elsaid2023backpropagation}. Similar to the literature results, CANTS-N's colonies allowed their RNNs to train for 30 epochs using data collected from measurements from 12 burners of a coal-fired power plant. The dataset is multivariate and non-seasonal, comprising 12 input variables, potentially dependent. The dataset was divided into a training set of 1875 steps and a test set of 625 steps (recorded per minute). 
Table~\ref{tab:mse} displays the average fitness of the best-performing RNNs, acquired from a series of experiments using ANTS, BP-CANTS, and BP-Free CANTS. These experiments involved different numbers of ants (agents) (ranging from 5 to 50 ants) and were conducted over 10-fold runs. The table also displays the average MSE of the best RNNs obtained from the evolving colonies of \method, demonstrating superior results ($mse_{average} = 0.000553$) compared to other methods.
The dynamic exchange of experience among colonies, guided by PSO, allowed them to adaptively explore the search space while converging to the best exploration strategy, ultimately achieving the best results.

\begin{figure}
    \centering
    \includegraphics[width=.48\textwidth]{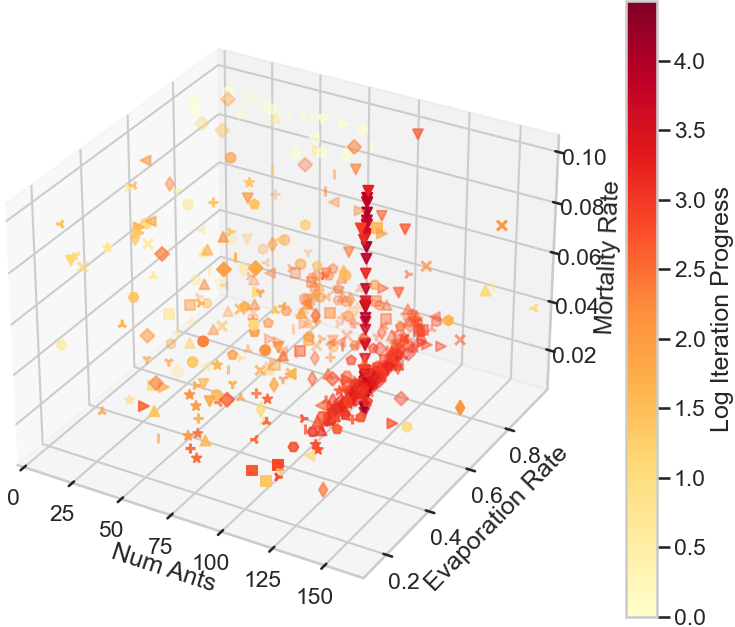}
    \caption{\centering \textit{\textbf{Colonies' Trajectories:} colonies' characteristics explore the colony traits search space using PSO through exploration \& exploitation}}
    \label{fig:pso}
\end{figure}

Figure~\ref{fig:pso} illustrates the trajectories of the traits of the twenty colonies over their 1000-generations lifetime in \method. Different marker shapes represent different illustrated colonies, while a logarithmic gradient colors the markers to reflect their proximity to the end of their lives. The figure showcases the initial dispersion of the colony characteristics and how they converge as they exchange experiences while interacting with their environment.

\begin{table}[h]
    \centering
    \begin{tabular}{r|c}
         \multicolumn{1}{l|}{\large\textbf{Method}} & \multicolumn{1}{c}{\large\textbf{Average Fitness (MSE)}} \\
                                                    & \multicolumn{1}{c}{(different colony size)} \\
        \midrule
         \textbf{ANTS}&  0.0308\\
         \textbf{BP-CANTS}&  0.0080\\
         \textbf{BP-free CANTS}&  0.00655\\
         \textbf{\method}&  \colorbox{lightgray}{0.0006}\\
         & 
    \end{tabular}
    \caption{\textbf{Average Fitness}}
    \label{tab:mse}
\end{table}

\section{Discussion \& Conclusion}
\label{sec:summary}


In this section, we amalgamate our discussions on the methodology, results, and their implications to offer a comprehensive view of CANTS-N and its significance in the realm of NAS and NE.

\paragraph{\textbf{Methodological Innovation:}}
In the relentless pursuit of optimizing neural architectures, CANTS-N stands as a testament to innovative methodological thinking. With no conventional optimization approach, CANTS-N orchestrates colonies of intelligent agents, each driven by a dynamic equilibrium between their insatiable hunger for exploration and their need for effective exploitation. These agents, akin to diligent foraging ants, dynamically interact within a continuous search space, navigating the intricate landscape of neural architectures.

CANTS-N's innovative approach is scalable and adaptable. The experiment's utilization of 201 CPU cores on a high-performance computing (HPC) cluster, each core running a separate MPI process, showcases its capabilities to harness the power of numerous processors efficiently, promising to challenge the norms of computational exploration.

\paragraph{\textbf{Outstanding Results:}}
The results obtained through CANTS-N's innovative methodology deserve attention. The experiment involved the creation of 20 colonies, each generating 1000 RNN structures, pushing the boundaries of NAS and NE research. These colonies, fueled by collective intelligence, engaged in strategic parameter exchanges, aiming for excellence. CANTS-N's dynamic approach, with colonies sharing experiences and adapting strategies, led to remarkable outcomes.

In a comparative analysis against established methods, including BP-free CANTS, BP-CANTS, and ANTS, CANTS-N outperformed them by an order of magnitude, demonstrating the potential of collaborative innovation in advancing NAS and NE research.

\paragraph{\textbf{Provoking New Horizons:}}
 While the agents in \method demonstrate proficiency as adept and intelligent learners~\cite{elsaid2023backpropagation}, they currently lack the cognitive capacity to assess the epistemic significance of the hidden states within their perceptual constructs of their world, actions, and proprioceptive channels~\cite{ueltzhoffer2018deep}. To bridge this perceptual gap and enhance their cognitive capabilities, we propose the future work of applying the concept of Active Inference (AI) and its free-energy principle~\cite{parr2022active} as a pivotal step. This enhancement is poised to not only empower agents in \method but also promises to advance the efficiency of Ant Colony Optimization in NAS and NE, as agents and probably colonies -- as myrmecologists like to consider them as the actual living and evolving organism, not the ants\cite{gordon2010ant} -- will be better equipped to perform their tasks with heightened cognitive acumen.

As part of our future research agenda, we envision the integration of AI techniques within the NAS framework. This innovative approach, grounded in the principles of AI and its free-energy concept, holds immense potential to revolutionize the field by infusing cognitive capabilities into the NAS process. By incorporating AI into NAS, we aim to unlock new horizons for achieving optimal neural architectures, propelling our research into the forefront of AI-driven NAS and NE methodologies.

\bibliographystyle{acm}
\bibliography{bib}  






\end{document}